\documentclass[conference,a4paper]{APSIPA2026}
\usepackage{amsmath}
\usepackage{graphicx}
\usepackage{multirow}
\usepackage{threeparttable}
\usepackage{booktabs}
\usepackage[backend=biber,style=ieee,]{biblatex}
\usepackage{amssymb}

\usepackage{array}
\usepackage{hyperref}
\usepackage{stfloats} 

\usepackage{geometry}
\usepackage{fancyhdr}

\fancypagestyle{firststyle}{
  \fancyhf{}
  \fancyhead[C]{2026 Asia Pacific Signal and Information Processing Association Annual Summit and Conference (APSIPA ASC)}
}

\begin{document}

\title{DispatchRAG: Grounding Emergency Dispatch Decisions in Real-World Protocols \\ from Traffic Accident Video}

\author{
\authorblockN{
    Muhammad Sulthan Adhipradhana,
    Ehsan Javanmardi,
    Naren Bao, and
    Manabu Tsukada
}
\authorblockA{
Graduate School of Information Science and Technology, University of Tokyo, Tokyo, Japan \\
E-mail: \{adhi, ejavanmardi, naren, mtsukada\}@g.ecc.u-tokyo.ac.jp}
Project page: \url{https://adhipradhana.github.io/DispatchRAG/}
}

\maketitle
\thispagestyle{firststyle}
\pagestyle{empty}

\begin{abstract}
  Assessing the severity of a traffic accident scenario is important to decide which emergency service to dispatch. 
  Missing an ambulance dispatch on a pedestrian accident is a fatal issue that can lead to death. Recently, Vision-Language Models (VLMs) 
  have been a promising tool for accident reasoning, yet many VLMs are not grounded in real-life accident response protocols, 
  making them not usable in accident severity assessment off-the-shelf. We introduced DispatchRAG, a decision-support framework grounded in real-life Japanese traffic-accident response protocols, designed to assist emergency dispatchers by using VLMs to generate an appropriate emergency response recommendation during an emergency scenario. Utilizing a RAG-based retrieval 
  mechanism to retrieve the most relevant accident protocol and an LLM-powered reasoner to suggest the most proper response. 
  To support evaluation, we introduce Accident Dispatch Dataset, a comprehensive dataset of accident assessment 
  and emergency response according to Japanese accident response protocols adapted from the MM-AU dataset \cite{fang2024mmau}. 
  We validate our framework on the Accident Dispatch Dataset, showing strong performance across various accident scenarios compared to the baseline VLM, pointing toward a decision-support tool that can help autonomous vehicles and human dispatchers respond faster to both their own and nearby accidents.
\end{abstract}


\section{Introduction}
In 2023, there were approximately 1.19 million traffic accident deaths \cite{world2024global}. 
The survivability rate of these victims depends on how quickly emergency services are dispatched. However, the average 
response time from the accident to a report call is around 4.61 minutes in urban areas and can go up to 7.24 minutes 
in rural areas \cite{wu2015updated}. This number is correlated with the fact that rural areas have fewer bystanders 
to report the accident compared to urban areas. With this gap, deploying an automatic decision-support tool
that can assess the damage severity and suggest the proper emergency response dispatch to human dispatchers is needed to
shorten the response time, especially in rural areas where no one may be present to call at all. Integrated into an autonomous vehicle, such a
supporting tool could help report not only the vehicle's own crash but also accidents
it witnesses involving other vehicles, turning every passing vehicle into a
potential reporter.

Recent work on traffic accident analysis~\cite{shi2025scvlm, xu2021sutd, sheng2025safeplug, gan2026crashsight} has 
begun to leverage Vision-Language Models (VLMs) to understand and classify traffic anomalies. However, these efforts 
largely stop at recognizing what happened. The closest to ours is EchoTraffic~\cite{xing2025echotraffic}, which moves toward 
response but only produces general recommendations. It neither assesses the severity of the accident nor decides 
which emergency units should be dispatched. As a result, no existing work provides decision support for the task that an emergency dispatcher
performs, assessing the urgency of an accident and dispatching the appropriate services accordingly. Moreover, this gap cannot
be closed by prompting an off-the-shelf VLM, which lacks knowledge of real emergency protocols. Such models answer only
from parametric memory, cannot justify their decisions against any authoritative standard, and are prone to
hallucination, which disqualifies them from an unsupervised, safety-critical role. Building a reliable decision-support tool for accident response
requires grounding the model's decisions in real emergency and traffic protocols, which is the approach that we take in this work.

To bridge this gap, we introduce \textbf{DispatchRAG}, a retrieval-augmented framework for traffic accident assessment 
and emergency dispatch decisions. Inspired by the regulation-grounded decision-making of DriveReg \cite{cai2026drivereg}, 
DispatchRAG grounds every decision in real emergency protocols rather than in the model's parametric memory. Given a dashcam clip, 
the framework (1)~converts the scene into a textual description, (2)~retrieves the relevant emergency protocols 
from a curated corpus of Japanese traffic law and EMS guidelines, (3)~assesses the severity of the accident 
against those protocols through a law-grounded urgency guidebook, and (4)~recommends actions intended towards 
three points of view (POVs): the dispatching agency, the on-scene civilian, and the arriving EMS crew.

To evaluate our framework, we further introduce the \textbf{Accident
Dispatch Dataset}, a new benchmark of \textbf{500 dashcam accident
clips} built on the videos of the MM-AU dataset \cite{fang2024mmau}. Each clip
is annotated manually, blind to any model output and justified rule by rule
against the accompanying protocol corpus---with the urgency tier, the
emergency units to dispatch (POV~1), the civilian instructions
(POV~2), and the crew briefing (POV~3), together with the set of
applicable protocol rules.

Our contributions are summarized as follows:
\begin{itemize}
    \item We propose \textbf{DispatchRAG}, the first framework to
    provide decision support for the emergency dispatcher's task on
    accident video: it retrieves the applicable emergency protocols,
    assesses scene severity according to them, and produces a
    three-POV dispatch plan grounded in the cited rules.
    \item We introduce the \textbf{Accident Dispatch Dataset}, a
    benchmark of 500 dashcam accident clips paired with a bilingual
    corpus of real Japanese traffic-law and EMS protocols, annotated
    with urgency tiers, dispatch units, and per-POV response plans.
    \item Experiments show that our decomposed, protocol-grounded
    design outperforms end-to-end frontier VLMs on the structured
    dispatch decisions: urgency, units to dispatch, and transport.
\end{itemize}

\section{Related Work}

\subsection{Vision-Language Model for Traffic Accident Understanding}
Recent work has begun to apply Vision-Language Models to analyze traffic accidents. Benchmark works \cite{fang2024mmau, kim2025vru, 
xu2021sutd, gan2026crashsight} are providing datasets to evaluate the quality of VLMs in understanding traffic accidents. 
Recent frameworks proposed classifying accidents type with narrative descriptions \cite{shi2025scvlm}, fine-grained spatial 
and temporal grounding of the accidents \cite{sheng2025safeplug}, and adding audio cues to enhance anomaly detection 
under adversarial conditions, such as nighttime or heavy snow \cite{xing2025echotraffic}.  Collectively, these systems only 
cover scene understanding and do not produce an actionable emergency-response decision.

\subsection{From Accident Understanding to Emergency Response}
A smaller set of research directions moves from accident understanding toward emergency response. GPT-4V evaluates VLM performance 
on various accident videos and finds that a prompted zero-shot VLM can describe a scene and propose emergency response, such as notifying medical, fire, 
and police services \cite{zhou2024gptv}. Another work combines a detector, a captioner, and a language model to produce incident reports with dispatch suggestions~\cite{ahmed2024detection}. Other work targets post-collision cause and liability analysis rather than dispatch \cite{wang2023accidentgpt}. 
In conclusion, these systems are focused more on a dispatch recommendation generated as free-form text without 
grounding the decision in any emergency-dispatch protocol or legal standard, leaving them unsuitable for a real-life accident dispatching system.

\subsection{Retrieval-Augmented and Protocol-Grounded Dispatch}
Two adjacent directions inform our approach. First, retrieval-augmented reasoning has been used to make safety-critical driving decisions 
trustworthy and auditable. DriveReg retrieves applicable traffic regulations and reasons over them to judge the legality of driving maneuvers~\cite{cai2026drivereg}. 
However, it addresses normal-driving compliance rather than accidents or dispatch. Second, automated emergency dispatch triage itself is well 
studied, but operates only on text. Recent LLM-powered systems infer urgency or priority from dispatch-call transcripts, emergency narratives, and ambulance requests~\cite{williams2024acuity,shekhar2025ambulance,li2026dispatchmas}. These methods prioritize a request already described by a human caller, 
rather than perceiving the accident itself. In contrast, our work makes the dispatch decision directly from visual evidence 
of the accident scene and grounds each decision in retrieved rules of real emergency and traffic protocols with an explicitly cited rule, 
producing an auditable dispatch decision that prior work does not provide.

\section{Proposed Method}
\label{sec:method}

\begin{figure*}[t]
  \centering
  \includegraphics[width=\textwidth]{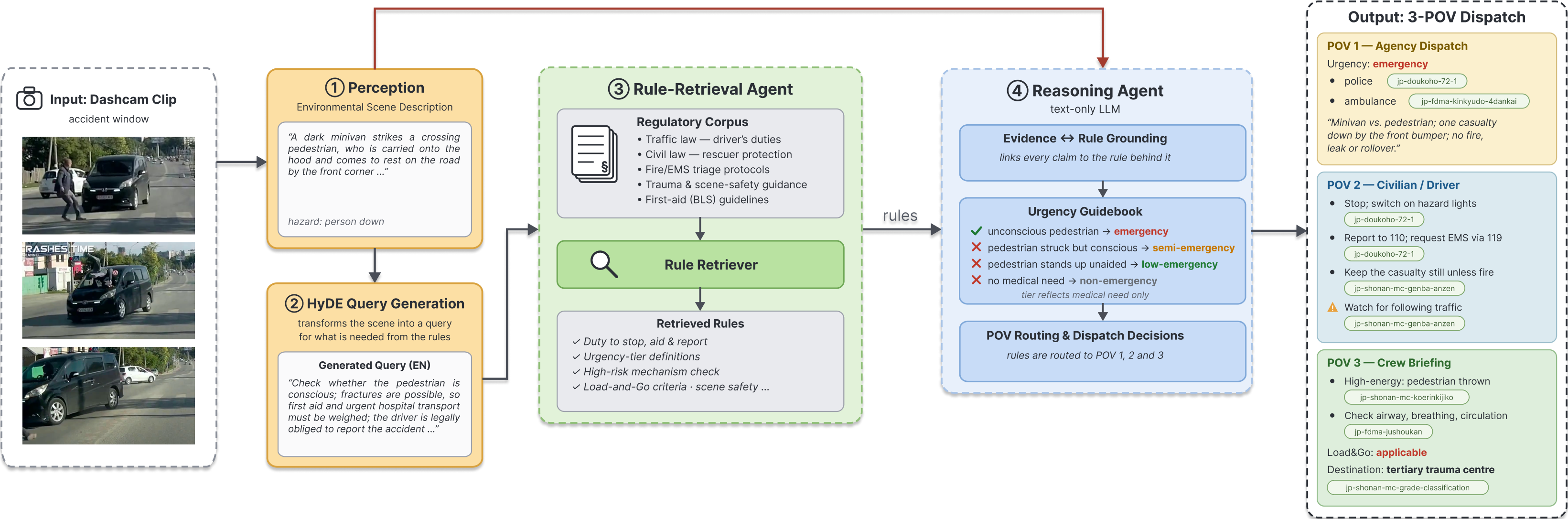}
  \caption{Overview of the proposed framework. A VLM converts the dashcam
  clip into a prose scene description, which is rewritten into
  an actionable retrieval query. The Rule-Retrieval
  Agent matches the query against a bilingual corpus of Japanese statutes
  and EMS protocols. The Reasoning Agent grounds each claim in observed
  evidence and a rule identifier, assigns the urgency tier via a
  law-grounded guidebook, and routes content to three audiences, producing a 3-POV dispatch plan.}
  \label{fig:pipeline}
\end{figure*}

\subsection{Overview}
\label{sec:overview}

Given a dashcam clip of a traffic accident, our goal is to
produce an emergency-dispatch plan in which every decision is grounded in
an authoritative Japanese law or medical protocol. End-to-end
VLMs can describe a crash, but they cannot state which rule
justified a decision. As shown in Fig.~\ref{fig:pipeline}, our framework
therefore decomposes the task into three stages: a \emph{Perception} stage
that turns the clip into a free-text scene description, a
\emph{Rule-Retrieval Agent} that fetches the applicable rules from a
curated regulatory corpus, and a \emph{Reasoning Agent} that composes a
three-perspective (3-POV) dispatch plan grounded in those rules. The 3-POV
plan targets the responding agency, the on-scene civilian, and the arriving
EMS crew. Each directive carries the cited rule that
justifies it. Video is consumed only at perception; all downstream
decisions operate on text and cited rule identifiers, so the plan can be
checked against its sources.

\subsection{Japanese Legal and Protocol Corpus}
\label{sec:corpus}

The corpus consists of 19 Japanese rule passages
curated from authoritative Japanese sources, which are 
the Road Traffic Act and Civil Code, the Fire and Disaster 
Management Agency's emergency service guidance (including the four-tier urgency classification), regional medical-control trauma protocols, 
and the Japan Resuscitation Council's basic life support
guidelines~\cite{doukoho, minpo, fdmaTriage, fdma119, shonanmc, jrc}. Combined, they cover 
the driver's legal duties, urgency assessment, unit dispatch, and prehospital care.

\subsection{Perception}
\label{sec:perception}

The perception VLM receives the 16 sampled accident frames and writes a 
natural-language paragraph describing the accident details. It includes collision type, vehicles involved, visible hazards (deformation, persons down, debris, vehicle rollover, fluid leak, fire), each vehicle's damage severity, and the scene environment. The prompt only
includes an observable scene description without including inferred information, such as 
clinical state. This is meant to keep the reasoning process from being biased by these keywords. 

\subsection{Rule-Retrieval Agent}
\label{sec:retrieval}

Since the VLM only describes the scene objectively, it does not contain
any actionable vocabulary that is stored in the rule corpus, which means 
a direct rule retrieval would miss important documents. Utilizing the HyDE paradigm
\cite{gao2023hyde}, an LLM rewrites the observation into a hypothetical
professional description of what this scene might require, such as the 
casualty's severity assessment, transport urgency, driver's legal duties, 
and scene-safety concerns. This information is inferred without citing
any article number or document titles. This enables the query to retrieve
action-oriented rules.

The query and the corpus passages are encoded by a multilingual
sentence-embedding model~\cite{multi2024bgem3}, and the agent retrieves the top-$k$
passages by cosine similarity to form the rule set passed to the
reasoner.

\subsection{Reasoning Agent} 
\label{sec:reasoner}
The Reasoning Agent is a text-only LLM that receives the scene description and the retrieved rules and writes the dispatch plan. Three mechanisms
make the output auditable and reproducible.

\subsubsection{Evidence--rule grounding}
The agent may cite only law identifiers from the retrieved rules, and every claim must be paired with the observed evidence that supports it; conditional
phrasing is reserved for genuinely unobservable facts (e.g., internal
injury).

\subsubsection{Law-grounded urgency guidebook}
The system maps observable cues onto the four-tier severity scale of the 119
emergency-call protocol, each tier carrying its dispatch implication. The
\emph{emergency} tier (Red) covers the most time-critical conditions:
suspected cardiopulmonary arrest, an unconscious struck pedestrian, and
high-energy mechanisms such as vehicle fire or entrapment that demand rapid
arrival and transport. The \emph{semi-emergency} tier (Yellow) covers
conditions still needing medical judgment with less time pressure, requiring
a hospital visit within about two hours. The \emph{low-emergency} tier
(Green) covers conditions outside the first two that still require
examination, and the \emph{non-emergency} tier (White) covers conditions
needing no medical attention. The tier reflects medical need only, and forms
an ordered scale (non-emergency $<$ low-emergency $<$ semi-emergency $<$
emergency); over- and under-triage denote predicting a tier above or below
the ground truth.

\subsubsection{POV routing}
The produced content is routed to three audiences with different needs:
\begin{itemize}
  \item \textbf{POV~1 --- Agency dispatch} (119/110 dispatcher):
  machine-actionable fields, such as the urgency tier, the emergency units to dispatch
  (police, ambulance, fire-rescue, doctor-car) each with a cited reason and an incident summary.
  \item \textbf{POV~2 --- On-scene civilian} (driver, bystanders): short
  imperative do/warn/don't directives executable without medical
  training: driver duties during an accident, first aid to give or withhold, and
  scene-safety warnings.
  \item \textbf{POV~3 --- Arriving EMS crew}: injury-mechanism summary,
  arrival checklist (consciousness assessment, spinal precaution,
  injury assessment), the Load-and-Go decision, and the medical service destination.
\end{itemize}

\subsection{Dispatch Dataset and Annotation}
\label{sec:annotation}

We construct the \textbf{Accident Dispatch Dataset} by annotating 500
dashcam accident clips from MM-AU~\cite{fang2024mmau} with the full 3-POV
dispatch plan. The clips span a range of accident types, including
vehicle--vehicle, vehicle--motorcycle, vehicle--bicycle,
vehicle--pedestrian, vehicle--object, and single-vehicle accidents. All
clips are annotated manually, blind to any model output, with every field
justified by the cited corpus rules. Each annotation collects four field
groups:

\begin{itemize}
  \item \textbf{Observation tags} --- the objective visible facts:
  collision type, involved vehicles, environment, scene
  hazards, damage severity, and the post-impact state of a struck
  vulnerable road user (unconscious/not
  moving, moving but unable to stand, stands up unaided, not visible).
  \item \textbf{Urgency tier} --- assigned with the same law-grounded
  guidebook used by the reasoner (Sec.~\ref{sec:reasoner}), keeping the
  label grounded in the protocol rather than guessed.
  \item \textbf{Structured 3-POV plan} --- the ground-truth dispatch in
  the same schema that the system produced, every item is cited.
  \item \textbf{Applicable rules} --- the rule identifiers that
  ground the clip's dispatch decisions.
\end{itemize}

Table~\ref{tab:datasets} positions our dataset against prior traffic-video
benchmarks. These existing datasets are focused on explaining
what happened. EchoTraffic provides audio-visual QA 
pairs for anomaly understanding~\cite{xing2025echotraffic}, MM-AU annotates
object boxes, accident time windows, and causes for abductive
reasoning~\cite{fang2024mmau}, SUTD-TrafficQA provides multiple-choice
questions across six event-reasoning tasks~\cite{xu2021sutd}, and
VRU-Accident and CrashSight pair accident clips with QA and
dense-caption annotations~\cite{kim2025vru, gan2026crashsight}. None of
them labels the severity assessment and emergency dispatch decision. While several of these datasets are larger, ours is the first to annotate each clip with an actionable
dispatch decision, a four-tier urgency label, the emergency units to send,
and per-POV instructions. Ours is also the first to ground every label in cited
real-world legal and medical protocols, making the annotations grounded
rather than subjective. 

\begin{table}[h]
\centering
\caption{The Accident Dispatch Dataset versus prior traffic-video
benchmarks. Urg.\ = urgency-tier label; Disp.\ = emergency-unit dispatch
label; Law = decisions grounded in cited legal/medical protocols.}
\label{tab:datasets}
\footnotesize
\setlength{\tabcolsep}{1.6pt}
\begin{tabular}{llcccc}
\toprule
Dataset & Annotation & \#Clips & Urg.\ & Disp.\ & Law \\
\midrule
EchoTraffic \cite{xing2025echotraffic} & audio--visual QA      & 29{,}865 & $\times$ & $\times$ & $\times$ \\
MM-AU \cite{fang2024mmau}           & boxes, time, cause       & 11{,}727 & $\times$ & $\times$ & $\times$ \\
SUTD-TrafficQA \cite{xu2021sutd}    & event-reason.\ QA        & 10{,}080 & $\times$ & $\times$ & $\times$ \\
VRU-Accident \cite{kim2025vru}      & VQA + caption            & 1{,}000  & $\times$ & $\times$ & $\times$ \\
CrashSight \cite{gan2026crashsight} & VQA + caption            & 250      & $\times$ & $\times$ & $\times$ \\
\textbf{Ours}                       & \textbf{3-POV dispatch}  & 500 & \checkmark & \checkmark & \checkmark \\
\bottomrule
\end{tabular}
\end{table}

\section{Experiments}
\label{sec:experiments}

\subsection{Experimental Setup}
\label{sec:setup}

\subsubsection{Dataset}
The Accident Dispatch Dataset consists of 500 annotated clips prepared under
the protocol of Sec.~\ref{sec:annotation}; we report results on 
the whole dataset. Scoring covers all ground-truth clips.

\subsubsection{Metrics}
For the structured decisions: \emph{Unit F1} on the dispatched unit set;
\emph{Urgency EM}, exact match on the four-tier label, and \emph{Urgency
Ord.}, an ordinal agreement score ($1-|g-p|/3$ over the ranked tiers) that
credits near-miss tiers; and \emph{Load-and-Go Accuracy} on the POV~3
transport decision. Because urgency is ordinal, we also report the
directional error rates \emph{over-triage} (predicting a tier above the
ground truth) and \emph{under-triage} (predicting below); a well-calibrated system minimizes both rather than trading one for the other. 

We separately analyze the binary ambulance decision (dispatch or not) with accuracy, precision, recall, and F1, since recall alone can be inflated by indiscriminately dispatching to every clip. Content quality uses an LLM
judge (per-POV \emph{coverage} of ground-truth directives). To avoid
single-judge and self-preference bias, every judge's score is the mean of a
three-judge panel drawn from three different model families (Gemma-2-27B,
Llama-3.3-70B, and Pixtral-12B), none of which is a system under test.

\subsubsection{Baseline models}
We compare four methods. \textbf{Framework (ours)} is the decomposed pipeline
of Sec.~\ref{sec:method}. The baselines are end-to-end \emph{VLMs with
Retrieved Rules}: each receives the sampled frames \emph{and the same
retrieved rules} and emits the 3-POV plan in one call, instantiated with
Qwen3-VL-32B\cite{bai2025qwen3vl}, GPT-5.4-mini, and Gemini-2.5-Flash\cite{comanici2025gemini}. Since all methods see the identical rule
context, the comparison isolates the perception/architecture split.

\subsubsection{Implementation details}
Perception uses GPT-5.4-mini over 16 uniformly sampled frames; the
retriever uses BGE-M3\cite{multi2024bgem3} dense embeddings with a fixed top-$k$ budget; the
reasoner is Qwen3-32B \cite{yang2025qwen3} with the 
urgency guidebook embedded in its system prompt.

\subsection{Main Results}
\label{sec:results}

\begin{table*}[t]
\centering
\caption{Structured dispatch decisions and LLM-judge content quality
($n = 500$). EM = exact match; Ord.\ = ordinal agreement, mean
$1-|g-p|/3$ over the ranked tiers; O-Tri/U-Tri = over-/under-triage rate;
coverage = recall of ground-truth directives per POV, averaged over a
three-judge panel (Gemma-2-27B, Llama-3.3-70B, Pixtral-12B)}
\label{tab:structured}
\footnotesize
\setlength{\tabcolsep}{6pt}
\begin{tabular}{lcccccc@{\hspace{20pt}}ccc}
\toprule
 & \multicolumn{6}{c}{Structured dispatch decisions} &
 \multicolumn{3}{c}{Judge coverage} \\
\cmidrule(lr){2-7}\cmidrule(lr){8-10}
 & Unit & \multicolumn{4}{c}{Urgency} & L\&G & & & \\
\cmidrule(lr){3-6}
Method & F1 $\uparrow$ & EM $\uparrow$ & Ord.\ $\uparrow$ & O-Tri $\downarrow$ & U-Tri $\downarrow$ & Acc $\uparrow$ & POV1 $\uparrow$ & POV2 $\uparrow$ & POV3 $\uparrow$ \\
\midrule
Qwen3-VL-32B + Rules      & 0.82 & 0.35 & 0.68 & 0.50 & \textbf{0.13} & 0.38 & 0.43 & 0.47 & 0.50 \\
GPT-5.4-mini + Rules      & 0.75 & 0.35 & 0.70 & 0.43 & 0.22 & 0.39 & \textbf{0.56} & \textbf{0.67} & \textbf{0.66} \\
Gemini-2.5 + Rules        & 0.80 & 0.23 & 0.63 & 0.60 & 0.17 & 0.32 & 0.43 & 0.52 & 0.51 \\
\textbf{Framework (ours)} & \textbf{0.85} & \textbf{0.47} & \textbf{0.77} & \textbf{0.28} & 0.25 & \textbf{0.55} & 0.55 & 0.56 & 0.58 \\
\bottomrule
\end{tabular}
\end{table*}

\subsubsection{Structured dispatch results}
On the structured decisions (left of Table~\ref{tab:structured}), the
framework leads every dispatcher-facing call: unit selection, urgency, and transport, compared to the end-to-end baselines. We explain each result
below.

\subsubsection{Urgency tier}
Urgency is the hardest field, since it decides every other decision. On
the exact-match accuracy (EM), the framework reaches 0.47 against 0.23--0.35 for the baselines, but an exact-match alone is harsh. Therefore, we add a comparison by providing an ordinal scale value, so an off-by-one call is a near miss rather than an outright error.
The ordinal score (Ord.\ 0.77 vs.\ 0.63--0.70), which gives a full score
for an exact tier and degrades linearly with tier distance, showing that even when the framework misses the exact tier, it lands closer on the scale. 

It is also the best calibrated: the framework over-triages on only 28\% of
clips versus 43--60\% for the baselines (O-Tri column). The baselines hold
under-triage low only by escalating almost everything, a non-discriminating
strategy whose cost surfaces in the ambulance analysis below
(Table~\ref{tab:ambulance}), where it yields up to three times more false
positives. The framework instead balances both error directions.

\subsubsection{Ambulance dispatch: calibration over recall}
\label{sec:ambulance_dispatch}
The most important call is whether to dispatch an ambulance at all.
Table~\ref{tab:ambulance} scores this binary over all 500 clips (320 require an ambulance, 180 do not). The framework's higher 
under-triage might seem unsafe compared to the baselines' near-perfect recall, but the confusion counts show why that recall is misleading: the baselines reach up to 0.99 only by dispatching in almost every clip. Gemini-2.5 predicts no ambulance on just 5 of 180
clips and produces 175 false positives, approximately three times the
framework's 58. Recall is trivially maximized by always escalating. Thus, the important objective should be calibration, in which the framework trades a little recall (0.81) for far higher precision (0.82), giving the best accuracy (0.76) and F1 (0.81) score. This implies the framework is the safest and best-balanced method compared to others that send an ambulance to every situation.

\begin{table}[t]
\centering
\caption{Ambulance-dispatch decision (binary, $n=500$; 320 require an
ambulance, 180 do not). TP/TN = correct dispatch/withhold;
FP = unnecessary dispatch (over-triage); FN = missed dispatch
(under-triage).}
\label{tab:ambulance}
\footnotesize
\setlength{\tabcolsep}{4pt}
\begin{tabular}{lcccccc}
\toprule
Method & Acc $\uparrow$ & Rec.\ $\uparrow$ & Prec.\ $\uparrow$ & F1 $\uparrow$ & TP/TN/FP/FN \\
\midrule
Qwen3-VL-32B + Rules & 0.72 & 0.92 & 0.72 & \textbf{0.81} & 295/65/115/25 \\
GPT-5.4-mini + Rules & 0.69 & 0.97 & 0.68 & 0.80 & 311/34/146/9 \\
Gemini-2.5 + Rules   & 0.64 & \textbf{0.99} & 0.64 & 0.78 & 316/5/175/4 \\
\textbf{Framework (ours)} & \textbf{0.76} & 0.81 & \textbf{0.82} & \textbf{0.81} & 258/122/58/62 \\
\bottomrule
\end{tabular}
\end{table}

\subsubsection{Emergency unit dispatch and transport}
On the set of dispatched units, the framework attains the highest Emergency Unit F1 (0.85 vs.\ 0.75--0.82). 
The transport decision is evaluated by Load-and-Go,
the POV3 prehospital call to transport the patient to the hospital
immediately, performing only life-saving interventions on the scene, rather
than treating in place, reserved for high-energy or time-critical injuries.
The framework makes this call best as well (L\&G Acc 0.55 vs.\ 0.32--0.39).

\subsubsection{Content quality}
On the prose quality (right of Table~\ref{tab:structured}), the result
separates cleanly. GPT-5.4-mini with retrieved rules writes the richest
directives, leading judge coverage across all three POVs, while the
framework's text is terser and more structured. The two are complementary: the framework wins the \emph{structured} dispatch, while a frontier VLM writes the richer \emph{narrative} crew and civilian content.

\begin{figure*}[t]
  \centering
  \includegraphics[width=0.8\textwidth]{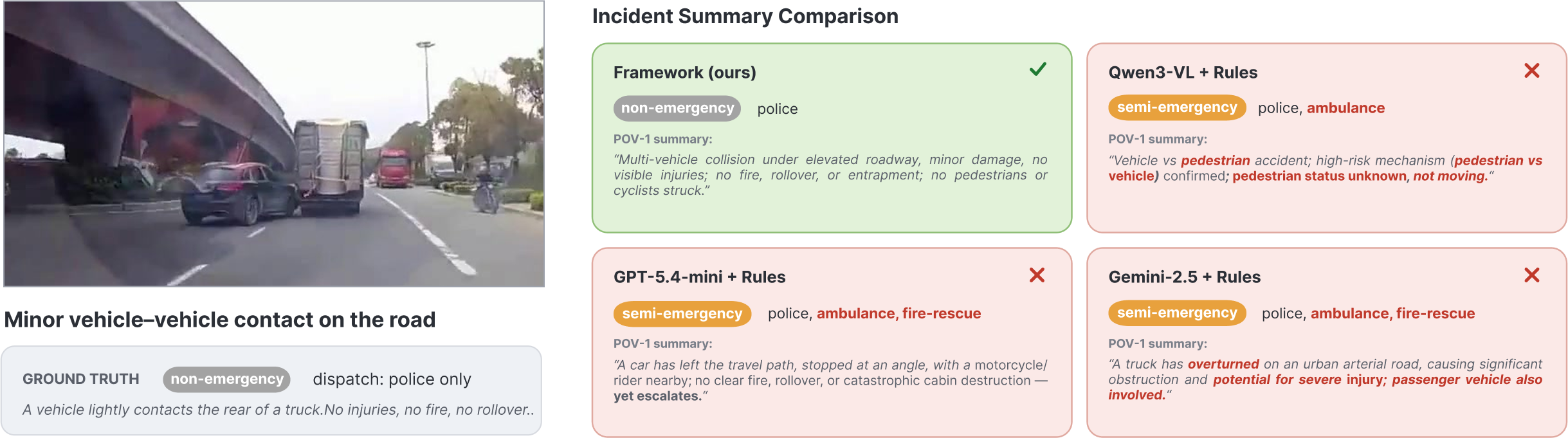}
  \caption{On a minor vehicle-vehicle contact (GT: \emph{non-emergency},
police only), the rule-augmented VLMs over-triage. Qwen3-VL-32B and Gemini-2.5 are hallucinating a severe element absent from the clip, while GPT-5.4-mini is escalating despite an accurate and benign description. It predicts unnecessary units. The framework holds the correct tier and dispatch.}
  \label{fig:qualbetter}
\end{figure*}

\begin{table}[ht]
\centering
\caption{Recall of the ground-truth cited
rules with perception and reasoner held fixed.
Only the query/scorer varied.}
\label{tab:retrieval}
\footnotesize
\setlength{\tabcolsep}{8pt}
\begin{tabular}{lc}
\toprule
Retrieval method & Recall $\uparrow$ \\
\midrule
BM25 \cite{robertson2009probabilistic} & 0.71 \\
Direct Observation & 0.73 \\
\textbf{HyDE (ours)} & \textbf{0.76} \\
\bottomrule
\end{tabular}
\end{table}

\subsubsection{Retrieval ablation}
Table~\ref{tab:retrieval} isolates the query formulation, holding
perception and the reasoner fixed and varying only how the corpus is
queried. The HyDE recalls the ground-truth cited rules best
(0.76), ahead of a query built directly from the raw observation
(0.73) and lexical BM25 (0.71). This suggests the HyDE paradigm transforms the query 
into an action-oriented register of the rule text, closing the vocabulary gap of Sec.~\ref{sec:retrieval}. 

\subsubsection{Limitations}
Two limitations remain. First, the framework trades recall for precision on
the ambulance call: it is the best-calibrated method overall, but a small fraction of true ambulance cases are under-triaged, and raising recall
without re-inflating false positives is an open problem. Second, perception
operates on a single forward dashcam view, so occluded or out-of-frame
casualties can be missed; a multi-view or sensor-fused input would relax
this constraint.

\subsection{Qualitative Comparison}
\label{sec:qualitative}

Fig.~\ref{fig:qualbetter} illustrates that the baselines over-triage a
non-emergency case of a minor car-to-truck contact. They fail in two ways:
Qwen3-VL-32B and Gemini-2.5 hallucinate a severe element absent from the
scene, a phantom pedestrian and an overturned truck, while
GPT-5.4-mini describes the scene accurately, correctly noting no fire or
rollover, yet still escalates and over-dispatches. In both modes, the
framework's rule-grounded tiering keeps the call grounded in what is
actually visible.

\section{Conclusion}
\label{sec:conclusion}

We presented DispatchRAG, a retrieval-augmented framework that turns a
dashcam accident clip into a law-grounded, three-perspective emergency
dispatch plan, together with the Accident Dispatch Dataset, the first
benchmark to annotate accident video with an actionable response, urgency
tier, dispatched units, and per-POV instructions grounded in cited
Japanese legal and medical protocols.

Our experiments support two conclusions. First, decomposition
delivers the most reliable critical decisions. Separating perception from
rule-grounded reasoning yields the most accurate urgency, unit, and
transport decisions and the best-calibrated triage, avoiding the
over-triage that escalates the end-to-end baselines, including newer
frontier VLMs running end-to-end. Second, the two methods are
complementary. Our framework produces an auditable, well-calibrated
structured plan with rule-level citations, while a frontier VLM writes
richer free-text sentences.

Several directions remain for future work. First, we plan to expand the
dataset with more annotated clips to cover a wider range of accident
scenarios. Second, we aim to extend the corpus beyond Japanese protocols to
the emergency and traffic regulations of other jurisdictions, broadening the
framework's applicability. Third, we intend to expand the framework in
a multi-vehicle setting, enabling collaborative perception that can assess damages from multiple views, addressing the occlusion limitation of a single camera viewpoint. 

\section{Acknowledgment}
This work was supported by JST, CRONOS, Japan Grant Number JPMJCS24K8.

\printbibliography

\end{document}